\pdfoutput=1

\documentclass[11pt]{article}

\usepackage[]{acl}

\usepackage{times}
\usepackage{latexsym}
\usepackage{booktabs}
\usepackage{multirow}
\usepackage[textsize=scriptsize]{todonotes} 
\usepackage{graphicx}
\usepackage{tabularx}
\usepackage{wrapfig}
\usepackage{multirow}
\usepackage{rotating}

\usepackage[inline]{enumitem}
\usepackage{tablefootnote}
\usepackage[T1]{fontenc}

\usepackage[utf8]{inputenc}

\newcommand{\geo}{\textsc{Geo}}
\newcommand{\rsq}{\textsc{Recv}}
\newcommand{\nat}{\textsc{Nature}}
\newcommand{\pen}{\textsc{Pen}}
\newcommand{\annot}{\textsc{Annot}}
\newcommand{\kw}{\textsc{Kw}}
\newcommand{\stdref}{\textsc{StdRef}}
\newcommand{\fk}{\textsc{Bert-FK}}
\newcommand{\aos}{\textsc{AOSumm}}

\usepackage{microtype}
\usepackage{enumitem}

\title{\textsc{AspectNews}: Aspect-Oriented Summarization of News Documents}

\author{Ojas Ahuja$^{1}$, Jiacheng Xu$^{1}$, Akshay Gupta$^{1}$, Kevin Horecka$^{2}$, Greg Durrett$^{1}$ \\
$^{1}$The University of Texas at Austin \\ $^{2}$Walmart NexTech\\ 
\texttt{\{ojas,jcxu\}@utexas.edu}, \texttt{gdurrett@cs.utexas.edu}}

\begin{document}
\maketitle
\begin{abstract}
Generic summaries try to cover an entire document and query-based summaries try to answer document-specific questions. But real users' needs often fall in between these extremes and correspond to aspects, high-level topics discussed among similar types of documents.
In this paper, we collect a dataset of realistic aspect-oriented summaries, \textsc{AspectNews}, which covers different subtopics about articles in news sub-domains. We annotate data across two domains of articles, earthquakes and fraud investigations, where each article is annotated with two distinct summaries focusing on different aspects for each domain. A system producing a single generic summary cannot concisely satisfy both aspects. Our focus in evaluation is how well existing techniques can generalize to these domains without seeing in-domain training data, so we turn to techniques to construct synthetic training data that have been used in query-focused summarization work. We compare several training schemes that differ in how strongly keywords are used and how oracle summaries are extracted. Our evaluation shows that our final approach yields (a) focused summaries, better than those from a generic summarization system or from keyword matching; (b) a system sensitive to the choice of keywords.\footnote{Code is available at \url{https://github.com/oja/aosumm}}
\end{abstract}

\section{Introduction}

Recent progress in text summarization \cite{see-etal-2017-get,liu-lapata-2019-text,pegasus,lewis-2019-bart} has been supported by the availability of large amounts of supervised data, such as the CNN/Daily Mail and XSum datasets \cite{hermann-2015-cnndm, xsum-emnlp}, which provide a single, generic, topic-agnostic summary. 
However, a document often contains different \textit{aspects} \cite{titov-mcdonald-2008-joint,woodsend-lapata-2012-multiple} that might be relevant to different users. For example, a political science researcher studying responses to earthquakes may want a summary with information about government-led recovery efforts and broader social impacts, not a high-level generic summary of what happened. Systems should be able to produce summaries tailored to the diverse information needs of different users. Crucially, these systems should be usable in realistic settings where a user is interested in vague \emph{aspects} of the document, instead of a highly focused query.

\begin{figure*}[t]
\centering
\includegraphics[width=\textwidth]{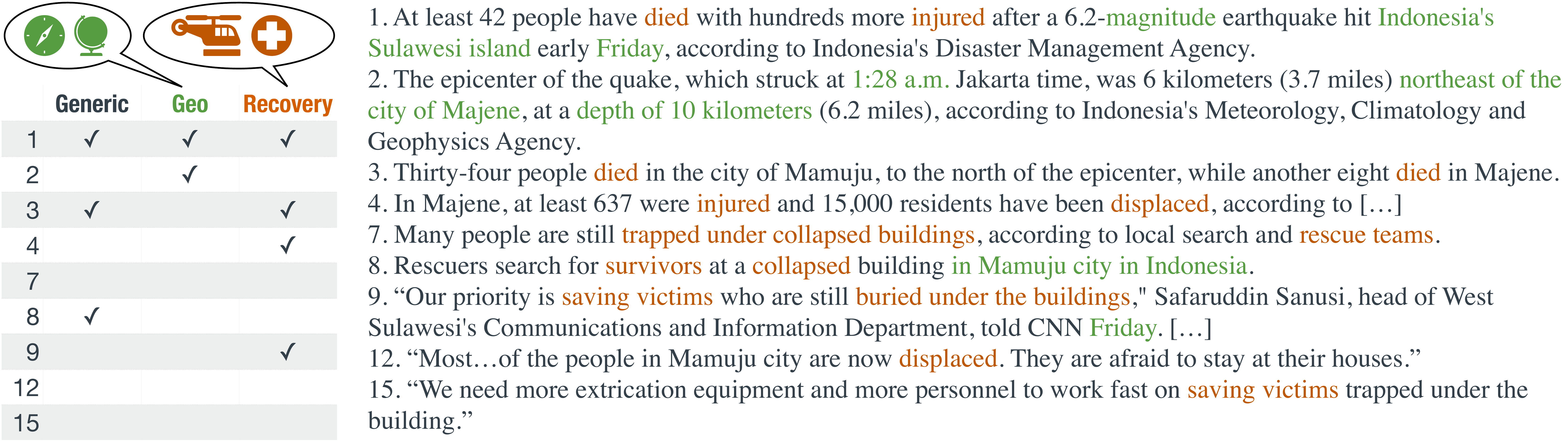}
\caption{Examples of an earthquake-related article paired with extractive summaries from the CNN/DM dataset. ``Generic'' represents the selection of a general purpose summarization model. ``Geo(graphy)'' (colored in green) and ``Recovery'' (colored in orange) indicate our aspects of interest for the summary. We highlight aspect-relevant phrases in the document.
}
    \label{fig:eq-docs}
\end{figure*}

In this work, we present a new dataset for evaluating single-document \emph{aspect-oriented} extractive summarization which we call \textsc{AspectNews}. We derive subsets of examples from CNN/Daily Mail following certain topics, namely earthquakes and fraud reports. These domains are special in that the articles within them have several aspects which are repeatedly mentioned across articles and form coherent topics, e.g., impact on human lives of an earthquake. We ask annotators to select sentences relevant to such information needs, which correspond to imagined use cases. Interannotator agreement on full summaries is low due to the inherent subjectivity of the task, so rather than coming up with a consensus summary, we instead primarily evaluate against soft labels based on the fraction of annotators selecting a given sentence.

To benchmark performance on this dataset, we build a system that can summarize a document conditioned on certain aspect-level keywords without assuming annotated training data for those aspects. Since there are no large-scale supervised training sets suitable for this purpose, we explore methods to generate aspect-oriented training data from generic summaries. We compare these with past approaches \citep{frermann-klementiev-2019-inducing} on their ability to adapt to our aspect-oriented setting, which requires taking aspectual keyword inputs (as opposed to specific entities or queries) and being appropriately sensitive to these keywords. 

Our experiments on our \textsc{AspectNews} dataset and the \textsc{SPACE} dataset \cite{Angelidis2021ExtractiveOS} find that our model produces summaries that score higher on agreement with human aspect-oriented annotations than generic summarization models, previous aspect-oriented models, and baselines such as keyword matching. Second, we find that the summaries our model generates are sensitive to the choice of keywords. Third, we find that our model performs competitively with leading models on the \textsc{SPACE} dataset in the multi-document setting. Finally, we find that abstractive query-focused systems \cite{he2020ctrlsum} hallucinate significantly in this setting, justifying our choice of an extractive framework here.

\section{Related Work}

Relatively little recent work has focused on aspect-oriented  summarization. One line of research focuses on summarization of documents with respect to specific queries \cite{baumel-etal-2014-query,krishna-srinivasan-2018-generating,frermann-klementiev-2019-inducing,he2020ctrlsum,xu-lapata-abstractive}. However, a query such as ``\emph{What facilities were damaged in the Oaxacan region?}'' is a document specific query, which cannot be applied to other earthquake news articles and bears more resemblance to the task of long-form question answering \cite{fan-etal-2019-eli5}. Our focus is closer to work on attribute extraction from opinions or reviews \cite{dong-etal-2017-learning,angelidis-lapata-2018-summarizing}, as factors like geographic details and recovery efforts are usually mentioned in many earthquake stories. 
Recent work has also begun to study summarization from an interactive perspective \cite{shapira-etal-2021-extending}; our approach could be naturally extended in this direction.

\paragraph{Methods} Historically, most work on query-focused summarization has addressed the multi-document setting. \citet{You2011ApplyingRM} apply regression models to this task, and \citet{Wei2008QuerysensitiveMR} approach the problem from the perspective of ranking sentences by their similarity to the query. These classic methods rely integrally on the multi-document setting, and so cannot be easily adapted to our setup. 
More recently, \citet{xu-lapata-2020-coarse} focus on multi-document summarization by modeling the applicability of candidate spans to both the query and their suitability in a summary. \citet{Angelidis2021ExtractiveOS} explore a method using quantized transformers for aspect-oriented summarization, which we compare to.

\paragraph{Datasets} There are several differences between \textsc{AspectNews} and other existing aspect-oriented summarization datasets. Firstly, \textsc{AspectNews} focuses on single-document summarization, while similar aspect-oriented datasets such as the \textsc{SPACE} dataset of reviews \cite{Angelidis2021ExtractiveOS} and other attribute extraction settings \cite{dong-etal-2017-learning,angelidis-lapata-2018-summarizing} are multi-document. Second, our dataset focuses on generalization \emph{to new aspect types}, rather than assuming we've trained on data with those same aspects; that is, how can we produce appropriate aspect-oriented summaries of earthquake articles even if we have not trained on any? Third, compared to query-focused settings, our aspect-oriented dataset is closer to the actual information needs of users, since users are often interested in summaries about broad subtopics rather than specific queries.

\begin{table*}[t]
\centering
\footnotesize
\setlength{\tabcolsep}{3pt}
\begin{tabular}{p{0.09\textwidth}  p{0.07\textwidth} | p{0.46\textwidth}|  p{0.33\textwidth}} 
\toprule
Domain    & Aspect       & Prompt                & Keywords                                   \\ \midrule
\multirow{2}{*}{Earthquake}   &  \geo & geography, region, or location       & region, location, country, geography, miles     \\
 &  \rsq & recovery and aid efforts (death toll and injuries, foreign/domestic government assistance, impact on survivors) & recovery, aid, survivor, injury, death\\ 
  \midrule
\multirow{2}{*}{Fraud}  
 &  \pen   & penalty or consequences for the fraudster, or for others & penalty, consequences, jailed, fined, court \\
& \nat       & nature of the fraud: the amount of money taken, benefits for the fraudster, and how the fraud worked   & amount, money, bank, stolen, time         \\

\bottomrule
\end{tabular}
\caption{Prompts and keywords used for each of our two domains: Earthquake and Fraud. These represent prominent topics that users might be interested in.}
\label{tab:annotation}
\end{table*}

The TAC 2010/2011 summarization datasets\footnote{\url{https://tac.nist.gov/2011/Summarization}} propose \textit{guided summarization} tasks that involve similar aspects. However, each article cluster in TAC has a single, fixed set of aspects that don't differ substantially from what a generic summary should capture. The DUC 2005/2006 task \cite{Dang2005OverviewOD} does not have aspects but rather can accept a ``granularity'' level at which to produce the summary. \citet{christensen-etal-2014-hierarchical} produce a hierarchy of relatively short summaries among multiple documents.

Other previous work \cite{he2020ctrlsum,xu-lapata-abstractive,tan-etal-2020-summarizing} proposes constructing keyword sets for each individual document for training. 
\citet{krishna-srinivasan-2018-generating,frermann-klementiev-2019-inducing} condition on topic tokens referring to the topic tags in metadata. Compared to these other approaches, we focus more on evaluation of aspects, as opposed to a purely keyword- and query-driven view.

\section{Aspect-Oriented Data Collection}

We begin by considering our target application: users who have specific information needs that they want to be satisfied. This consideration broadly falls under the category of \textbf{purpose factors} defined by \citet{Jones98automaticsummarising} and should be accounted for in the summarization process. 

Our data collection process involves the following steps: (1) Identifying clusters of articles in our \textbf{target domains} from a large corpus of news summaries. (2) Manually specifying multiple \textbf{user intents} per target domain, representing the \textit{aspect} of the summarization process. (3) \textbf{Crowdsourcing} annotation of extractive summaries in these domains based on the user intents.

\subsection{Target Domains}

We draw our datasets from the English-language CNN/Daily Mail summarization dataset \cite{hermann-2015-cnndm}. We manually identified two domains, \emph{earthquakes} and \emph{fraud}, based on inspecting clusters of articles in these domains. These two domains are ideal for two reasons. First, they contain a significant number of on-topic articles (over 200) after careful filtering. Second, the articles in these domains are reasonably homogeneous: each article would often feature at least broadly similar information about an event, making aspect-based summarization well-defined in these cases.\footnote{By contrast, other domains like legislation were too heterogeneous: articles about passing a bill may focus on different aspects of a bill's journey, comments or quotes by elected officials, impact of the legislation, or other factors. We could not come up with a plausible unified information need for the sorts of articles available in this dataset, although our eventual system can be applied to such documents if given appropriate guidance.} Although not completely universal, most earthquake articles refer to some information about each of two aspects here: \emph{geography} (\geo) and \emph{recovery} (\rsq).
Figure~\ref{fig:eq-docs} shows an example of an earthquake-related article. Similarly, most fraud articles include information about the \emph{penalty} (\pen) imposed for the fraud, and the \emph{nature} (\nat) of the fraud. 

To retrieve our examples from these two domains, we first encode each article in CNN/DM corpus $\mathcal{C}$ with a text encoder $E$. We adopt the Universal Sentence Encoder \cite{Cer2018UniversalSE} for its efficiency and robustness. We create an exemplar sentence for each domain to serve as the target to retrieve the most relevant content. We describe the choice of exemplar sentences in Section~\ref{app:exemp}. 
We measure the similarity of each candidate article $c$ and the exemplar sentence $s$ as the average of the cosine similarity between each of the candidate article's sentences $c_i$ and the exemplar, $sim(c, s) = \frac{1}{n} \sum_{i=1}^{n} \cos ( E(c_i), E(s))$.

We found this procedure to be more robust than simple keyword matching for retrieving articles with coherent aspects; for example, keyword matching for ``earthquakes'' resulted in returning articles primarily about tsunamis due to the imbalanced data distribution.

\subsection{Specifying User Intents}
With these two domains, we examine our dataset to derive aspects that simulate realistic information needs of users. 

Table~\ref{tab:annotation} describes the domain, aspect, annotation prompt and keywords used for evaluation. For each domain, we establish two aspects. Each aspect must be well-represented in the corpus and easy to understand by both readers and annotators. 
The authors annotated these aspects based on inspection of the articles and brainstorming about user intents based on scenarios. For example, the \emph{penalty} scenario was motivated by a real use case derived from the authors' colleagues investigating reporting of wrongdoing in news articles at scale, where summarization can be used to triage information.

\subsection{Crowdsourcing}

Finally, to construct actual extractive summaries for evaluation in these domains, we presented the user intents to annotators on Amazon Mechanical Turk. An annotator is shown a description of intent from Table~\ref{tab:annotation} along with an article and is asked to identify a few sentences from the article that constitute a summary. They can rate each sentence on a scale from 0 to 3 to account for some sentences being more relevant than others. Their final summary, which they are shown to confirm before submitting, consists of all sentences rated with a score of at least 1. The exact prompt is shown in the Appendix.

Each article was truncated to 10 sentences for ease of annotation. This assumption was reasonable for the two domains we considered, and the truncation approach has been used in \citet{see-etal-2017-get} without much performance degradation. We found that annotators were unlikely to read a full length article due to the inherent lead bias in news articles, so this also helped simplify the task. In order to maintain a high quality of annotations, we discard annotations that do not have at least a single selected sentence in common with at least a single other annotator on that sample. In practice, this only discards a handful of isolated annotations.

\begin{table}[t]
\footnotesize
\centering
\begin{tabular}{@{}r|ccc@{}}
\toprule
        & \# articles &  \# sent &  \# words  \\ \midrule
 {\pen} & 100 & 2.90 & 30.5 \\
 {\nat} & 100 & 2.79 & 29.9 \\ \midrule
 {\geo} & 100 & 2.53 & 28.4 \\
 {\rsq} & 100 & 2.76 & 27.0 \\
\bottomrule
\end{tabular}
\caption{Statistics for the collected datasets. For each aspect we collect 100 articles and each article is annotated by 5 Turkers. \#sent and \#words are the average number of sentences selected and average number of words in each sentence. }
\label{tab:dataset_stats}
\end{table}

\begin{table}[t]
\footnotesize
\centering
\begin{tabular}{@{}r|cccccc@{}}
\toprule
Agreement & 1 & 2 &3 &4 & 5 \\ \midrule
Freq (\%) & 19.61 &	29.26 &	25.16&	19.16&	6.80 \\
\bottomrule
\end{tabular}
\caption{Majority agreement distribution of 5 annotators on filtered collected data.}
\label{tab:sentence_agreement}
\end{table}

\begin{table}[t]
\footnotesize
\centering
\begin{tabular}{@{}r|cc@{}}
\toprule
{\stdref} vs.                   & Jaccard Sim. &  EM (\%) \\ \midrule
 {\pen} & 0.247 & 1.0 \\
 {\nat} & 0.249 & 2.0 \\
\midrule
 {\geo} & 0.265 & 2.0 \\
 {\rsq} & 0.201 & 1.0 \\
\bottomrule
\end{tabular}

\caption{Comparison of annotation labels and the non-query focused extractive oracle derived from reference summaries. We take the top-3 most common selected sentences from each aspect-oriented dataset and compute Jaccard similarity between the sets and the percentage of exact matches (EM).}
\label{tab:dataset_difference}
\end{table}

\subsection{Data Analysis \& Annotator Agreement}
In Table~\ref{tab:dataset_stats}, we show the basic statistics of the collected dataset. 
We show the distribution of the number of sentences agreed upon by the annotators in Table~\ref{tab:sentence_agreement}. We see that annotators somewhat agree in most cases, but relatively few sentences are uniformly agreed upon by all annotators. Our initial pilot studies also showed that annotators are often unsure where the cutoff is for information to be notable enough to include in a summary. We therefore view this disagreement as inherent to the task, and preserve these disagreements in evaluation rather than computing a consensus summary.

We also compare the overlap between aspect-oriented annotation and generic extractive oracle derived from reference summaries from CNN/DM. In Table~\ref{tab:dataset_difference}, the similarity and exact match\footnote{The number of annotated examples for each aspect is 100, so the EM is an integer.} between generic oracle summaries and the top 3 annotated sentences are fairly low, which means the annotated aspect driven summaries significantly differ from the standard extractive oracle.

\section{Building an Aspect-Oriented System}
\label{sec:model}
Our aspect-oriented data collection works well to create labeled evaluation data, but it is difficult to scale to produce a large training set. Identifying suitable domains and specifying user intents requires significant human effort, and collecting real test cases at scale would require a more involved user study. 

We build an aspect-oriented model without gold-labeled aspect-oriented training data. We do this by generating keywords for each article in CNN/DM, and training the model to learn the relationship between these keywords and a summary. Our system follows broadly similar principles to \citet{he2020ctrlsum}, but in an extractive setting. 

\subsection{Keyword-controlled Data}

We present a scheme to generate keywords for each document from the original dataset. CNN/DM consists of pairs $(D,S)$ of a document $D$ and associated summary $S$. We aim to augment these to form $(D,K,S')$ triples with keywords $K$ and a possibly modified summary $S'$. Our mixed augmentation technique requires training the model on \textbf{both} $(D,S)$ and $(D,K,S')$ for a given document.
We now describe the steps to create this data.

\definecolor{countries}{rgb}{0.00392156862745098,0.45098039215686275,0.6980392156862745}
\definecolor{budget}{rgb}{0.00784313725490196,0.6196078431372549,0.45098039215686275}
\definecolor{development}{rgb}{0.,0.,0.6}
\definecolor{10billion}{rgb}{0.8,0.47058823529411764,0.7372549019607844}
\definecolor{Turkey}{rgb}{0.792156862745098,0.5686274509803921,0.3803921568627451}

\begin{table}[t]
\centering
\scriptsize
\begin{tabular}{@{}p{7.4cm}@{}} \toprule
Article: 1. Justine Greening has called for a major shake-up in the EU aid \textcolor{budget}{budget} – as it emerged more than half the cash is squandered on relatively rich \textcolor{countries}{countries}. \\

2. The International \textcolor{development}{Development} Secretary challenged the basis of the £\textcolor{10billion}{10-billion}-a-year \textcolor{budget}{budget}, which channels cash to \textcolor{countries}{countries} such as \textcolor{Turkey}{Turkey}, Iceland and Brazil. \\

3. She is pressing for a major shift in policy to target resources at the poorest \textcolor{countries}{countries}. \\

4. International \textcolor{development}{Development} Secretary Justine Greening today insisted aid money [...] \\

5. Miss Greening held talks with ministers from [...]\\

7. Miss Greening said: ‘I don’t think it’s right that the EU still gives money to those \textcolor{countries}{countries} higher up the  [...]\\

9. Her intervention comes amid mounting concern about the EU aid \textcolor{budget}{budget}, which [...] total aid \textcolor{budget}{budget}.  [...]
 \\ \midrule
 Keywords: \textcolor{countries}{countries}, \textcolor{budget}{budget}, \textcolor{development}{development}, \textcolor{10billion}{10-billion}, \textcolor{Turkey}{Turkey} \\
 \bottomrule 
\end{tabular}
\caption{An example article from CNN/DM and keywords extracted. These keywords indicate both highly specific concepts and broad topic, but a model trained on data with appropriate reference summaries can learn to leverage either specific \emph{or generic} keywords in the summarization process. }
\label{fig:main}
\end{table}

\paragraph{Keyword Extraction} For each document in CNN/DM, we calculate the most important tokens in that document according to their TF-IDF ranking with respect to the entire corpus. Of these tokens, we select the ones that are present in the reference summary. This process selects tokens that are more likely to be consequential in affecting the output summary.

\paragraph{Reference Summary Computation} Since CNN/DM reference summaries are abstractive,  we need to derive extractive oracle summaries for training; these consist of sentence-level binary decisions $\mathbf{E} = E_1,\ldots,E_m$ for each sentence. Traditionally, this is done by finding a set of sentences that maximize ROUGE-2 (R2) with respect to the reference: $\textrm{argmax}_\mathbf{E} R2(\mathbf{E},S)$ \citep{gillick-favre-2009-scalable,Nallapati2017}. However, training the model to predict $P(S_1,\ldots,S_m \mid D, k)$, an extractive analogue of \citet{he2020ctrlsum}, was insufficient for our extractive model to learn to be sensitive to keywords; it merely learned to return a good generic summary regardless of what keywords were given.

To instill stronger dependence on the keywords, we made two modifications to this process. First, we modified the reference summary by concatenating the keywords with the reference summary \emph{before} computing the extractive oracle summary. This concatenation makes the oracle extraction more likely to select sentences containing the keywords, though modifying the reference summary requires maintaining a balance between the influence of keywords and of the original gold summary.

Second, we use BERTScore \cite[BS]{zhang-bertscore} rather than ROUGE-2 to identify sentences that closely match the reference summary. BERTScore turns out to boost the evaluation performance by a large margin, as shown in Table~\ref{tab:bertscore_ablation}, so we use BERTScore for oracle extraction for all our experiments. One reason for this is that the ROUGE-2 summaries favor exact keyword matches in selecting sentences, so the trained model simply learned to keyword matching in extreme cases. Our final reference summary is therefore $\textrm{argmax}_\mathbf{E} BS(\mathbf{E},S+nK)$, where $n$ is a hyperparameter we discuss next.

\paragraph{Keyword Intensity} To compute $n$, we introduce another parameter $r$ that controls the ratio of keyword tokens to original reference summary tokens. 
Higher values of $r$ lead to extracting sentences in a manner more closely approximating keyword matching, but yielding poor standalone summaries. On the other hand, lower values of $r$ may lead to generic summaries insensitive to the keywords. 
In practice, the number of times a keyword $w$ is concatenated to the original summary $S$ is defined as $n = r \times\frac{ \textrm{len}(S)}{\#(\textrm{keywords})}$ where $\textrm{len}(S)$ is the number of tokens in the original summaries and $\#(\textrm{keywords})$ is the total number of keywords available. When $r=1$, the  concatenated keywords have the same length of the original summary. 

\paragraph{Mixed Training} We explore a variant of training where we include training data with multiple variants of each original document from the dataset. Each document in the original dataset is mapped to two training samples, (1) a document without keywords and an unmodified oracle extractive summary, (2) a document with keywords and an oracle extractive summary using our modification procedure. 

\subsection{Aspect-Oriented Model}

Our model is trained to predict a summary $S$ from a document-keywords pair $(D,K)$. Following \textsc{BertSum} \cite{liu-lapata-2019-text}, we fine-tune BERT \cite{devlin2019bert} for extractive summarization using our modified CNN/Daily Mail dataset with keywords. During training, we prepend a special token followed by the keywords to the original document, and use the modified oracle extractive summary as the gold outputs. During inference, the keywords are user-defined. This scheme is similar to \citet{he2020ctrlsum}, but differs in that it is extractive.

We refer to this model, trained on our BERTScore references with the mixed training scheme, as \aos{}.


\begin{table*}[t]
\small
\centering
\setlength{\tabcolsep}{4pt}
\begin{tabular}{rrrrrrrrrrrrrrrrr}
\toprule
\multirow{2}{*}{Model}
& \multicolumn{4}{c}{\pen \annot} & \multicolumn{4}{c}{\nat \annot} & \multicolumn{4}{c}{\geo \annot} & \multicolumn{4}{c}{\rsq \annot}  \\
\cmidrule(lr){2-5} \cmidrule(lr){6-9} \cmidrule(lr){10-13} \cmidrule(lr){14-17}
& F$_{1}$ &  R-1 & R-2 & R-L & F$_{1}$ &  R-1 & R-2 & R-L & F$_{1}$ & R-1 & R-2 & R-L & F$_{1}$ &  R-1 & R-2 & R-L \\
\midrule
{\stdref} & 32.9     &  51.7   & 39.5  & 40.7 & 33.5    &  53.0  &  41.3  &  42.0  & 34.9    & 51.9  &  41.3 &   42.1 & 28.2  &     45.7 &   33.0  &  37.4\\
\textsc{Keyword} & 39.2    &  62.0  &  50.6  &  47.1 & 38.3   & 58.7  &  46.6  &  45.0 & 50.9 & 67.9  &  59.9  &  53.7 & 32.8  &     53.3  &  41.6  &  43.9\\
\textsc{QA} & 30.7   &    46.9   & 36.8 &   37.7 & 26.5   &   39.1   & 28.8  &  32.2 & \textbf{52.4}   & 63.0  &  58.9  &  \textbf{56.8} & 32.9      &  46.6   & 36.5  &  38.5\\
\midrule
\textsc{\textsc{BertSum}}   & 40.1   &  60.1 &   47.8 &   46.5 &  41.6  &   63.5  &  51.7  &  \textbf{49.4} & 46.4   & 65.4  &  56.4   & 51.4 & 37.3  &  55.8  &  44.8   & 44.6\\
{\fk} & 24.5  &  43.9  &  28.9  &  33.2 & 21.0 &   40.8  &  23.4   & 28.3 & 23.9  &  42.4  &  30.3  &  32.9 & 21.4  &  35.4  &  21.3   & 26.9 \\
\textsc{CTRLSum}  & N/A    &   47.8 &   30.2 &   33.0 & N/A &  51.7   & 35.3  &  35.4  & N/A  &  21.6  &  8.0  &  19.6 & N/A   & 32.3   & 11.6  &  19.2\\
{\aos} & \textbf{44.8}   &  \textbf{64.2} &   \textbf{54.1}  &  \textbf{51.6} & \textbf{45.2}   &  \textbf{64.4}  &  \textbf{53.9} &   48.0 & 49.9   & \textbf{69.1}  &  \textbf{61.2}   & 54.2  & \textbf{39.6}  &  \textbf{59.5}   & \textbf{49.1} &   \textbf{46.7}\\
\midrule
Max & 60.3 &  & & & 61.5 & & & & 70.2 & & & & 61.4 & & &  \\
\bottomrule
\end{tabular}
\caption{Performance comparison of our model ({\aos}) versus baselines on the \textsc{AspectNews} dataset in both the earthquakes and fraud domains, using our geography (\geo\annot) and recovery (\rsq\annot) aspects for the former and penalty (\pen\annot), and nature (\nat\annot) aspects for the latter. The last row displays the maximum possible F$_1$ score due to the disagreement of annotation. }
\label{tab:performance}
\end{table*}

\begin{table}[t]
\footnotesize
\centering
\begin{tabular}{@{}lrrrrrr@{}}
\toprule
                      & \multicolumn{1}{l}{\rotatebox[origin=c]{90}{Service}} & \multicolumn{1}{l}{\rotatebox[origin=c]{90}{Location}} & \multicolumn{1}{l}{\rotatebox[origin=c]{90}{Food}} & \multicolumn{1}{l}{\rotatebox[origin=c]{90}{Building}} & \multicolumn{1}{l}{\rotatebox[origin=c]{90}{Cleanliness}} & \multicolumn{1}{l}{\rotatebox[origin=c]{90}{Rooms}}     \\ \midrule
\textsc{BertSum}               & 12.4                       & 16.7                        & 13.0                    & 15.6                        & 13.8                           & 12.5 \\
\textsc{CTRLSum}               & 20.1                       & 18.6                        & 17.4                    & 18.9                        & 23.3                           & 19.7                         \\
\textsc{QT} & 26.0                       & 23.6                        & 17.7                    & 16.0                        & 25.1                           & 21.6                         \\ \midrule
\aos{}                & 26.9                       & 20.3                        & 17.4                    & 16.4                        & 22.8                           & 21.6                         \\ \bottomrule

\end{tabular}

\caption{ROUGE-L scores on the \textsc{SPACE} dataset of our model, \aos, versus \textsc{BertSum}, \textsc{CTRLSum}, and quantized transformer (\textsc{QT}). Despite being an extractive model, our approach is competitive with strong query-focused or aspect-based models.}
\label{tab:space_results}
\end{table}

\section{Experiments}
\label{sec:experiments}

We evaluate our model on the \textsc{AspectNews} dataset, comparing  performance on aspect-oriented summarization to several baselines. We additionally experiment on the \textsc{SPACE} multi-document dataset \cite{Angelidis2021ExtractiveOS} to provide a point of comparison on a prior dataset and show that our aspect-oriented method is competitive with other systems.

\subsection{Metrics}

On \textsc{AspectNews}, we evaluate our model against the annotations using using F$_1$ score and ROUGE scores. 
It is impossible to achieve 100 F$_1$ on this task due to inherent disagreement between annotators.
One downside of F$_1$ is that the model may be penalized even when the predicted sentence is very similar to the annotation, for this reason we also calculate ROUGE-1, -2, and -L scores \cite{lin-2004-rouge}. On the \textsc{SPACE} dataset, the gold summaries are abstractive, so we only calculate ROUGE scores.

\subsection{Baselines \& Competitor Models}

On the \textsc{SPACE} corpus, we primarily focus on comparisons to quantized transformer (QT) \cite{Angelidis2021ExtractiveOS} and \textsc{CTRLSum} \cite{he2020ctrlsum}. For the \textsc{AspectNews} dataset, we benchmark our system against several other models and baselines which we now describe.

\paragraph{Heuristic and QA Baselines} \textsc{Keyword} takes the keywords described in Table~\ref{tab:annotation} and greedily finds the first occurrence of each keyword in the input document. 
{\stdref} stands for the extractive oracle given the original reference summaries from CNN/DM. 
{\textsc{QA}} uses an ELMo-BiDAF question answering model \cite{Seo2017BidirectionalAF,peters-etal-2018-deep} to find answers to synthetic questions ``\emph{What is \{keyword\}?}'' for each keyword in the article. We select the sentence where the selected span is located as a sentence to extract.
Each of these three technique is an extractive baseline where top sentences are selected.

\paragraph{Summarization Baselines} We also compare our {\aos} model against text summarization models, and query-focused models from previous work (retrained or off-the-shelf). 
\begin{enumerate*}[label=(\roman*)]
  \item \textsc{BertSum} is a \texttt{bert-base-cased} extractive summarization model fine-tuned on CNN/DM \cite{liu-lapata-2019-text}. 
  \item {\fk} shares the similar model architecture as  \textsc{BertSum} but the training data comes from \citet{frermann-klementiev-2019-inducing}. This data is constructed by interleaving several articles from the CNN/DM dataset together, extracting a coarse aspect from the original URL of one of the article, and setting the new gold summary to match that article.
  \item \textsc{CTRLSum}
  is an off-the-shelf \textbf{abstractive} summarization model with the capability of conditioning on certain queries or prompts \cite{he2020ctrlsum}. 
  \item Our model {\aos} is based on \textsc{BertSum} and trained with techniques described in Section~\ref{sec:model}.
\end{enumerate*}

\subsection{Results}


\paragraph{\textsc{AspectNews}} The experimental results on \textsc{AspectNews} are shown in Table \ref{tab:performance}. We find that our model outperforms our baselines across F$_1$, ROUGE-1, ROUGE-2, and ROUGE-L scores. Significantly, our model generally outperforms keyword matching, demonstrating that semantic match information from training with the BERTScore oracle may be more useful than training with a ROUGE oracle in terms of reproducing annotators' judgments; recall that our model has not been trained on any \textsc{AspectNews} data and only on our synthetic data.

We note that our model's performance falls behind keyword matching some baselines in the geography aspect; this may be because the aspect is relatively homogeneous and can be easily approximated by keyword matching.

\paragraph{\textsc{SPACE}} The results on all the aspects of the \textsc{SPACE} dataset are shown in Table \ref{tab:space_results}. All of the aspect-oriented models exceed the performance of the generic summaries produced by \textsc{BertSum}. We also find that our model performs competitively with the quantized transformer (\textsc{QT}) \cite{Angelidis2021ExtractiveOS} and \textsc{CTRLSum} \cite{he2020ctrlsum} methods in this dataset. This is a surprising result: the {\aos} model is trained \emph{only} with out-of-domain synthetic data, without access to the aspects prior to keywords specified at test time. Additionally, this is an \emph{abstractive} task that we are applying an \emph{extractive} model to.

\begin{table}[t]
\centering
\footnotesize
\setlength{\tabcolsep}{3pt}
\begin{tabular}{@{}r|cccc|cccc@{}}
\toprule
\multicolumn{1}{r|}{ \kw  } & F$_{1}$ & R-1   & R-2   & R-L   & F$_{1}$ & R-1   & R-2   & R-L   \\ \midrule
                     & \multicolumn{4}{c|}{\pen \annot}     & \multicolumn{4}{c}{\nat \annot}        \\
\pen           & \textbf{44.8}  &   64.2 &   54.1   & \textbf{51.6} & 41.8   &   60.8  &  49.5  &  46.5 \\
\nat           & 44.3   &  \textbf{65.5}  &  \textbf{56.0}  &  51.3  &  \textbf{45.2}   &  \textbf{64.4} &   \textbf{53.9}   &  \textbf{48.0} \\
\midrule
 & \multicolumn{4}{c|}{\geo \annot}     & \multicolumn{4}{c}{\rsq \annot}     \\
\geo              & \textbf{49.9}  &  \textbf{69.1}  &  \textbf{61.2}  &  \textbf{54.2} & 38.0  &  56.2  &  45.3   & 46.2   \\
\rsq             & 42.8 & 60.4 &   49.7 &   47.8 & \textbf{39.6}  &  \textbf{59.5} &   \textbf{49.1}   & \textbf{46.7} \\ \bottomrule
\end{tabular}%
\caption{Keyword sensitivity analysis broken down by domain of \textsc{AspectNews}. }
\label{tab:kw-sens-domain}
\end{table}

\begin{table}[]
\footnotesize
\centering
\begin{tabular}{@{}r|cc@{}}
\toprule
                   & Jaccard Sim. &  EM (\%) \\ \midrule
{\pen \kw} vs. \nat \kw & 0.657         & 21.0            \\
{\geo \kw} vs. \rsq \kw     & 0.559         & 22.0             \\
\bottomrule
\end{tabular}
\caption{Difference in {\aos} outputs with different keywords. We compute Jaccard similarity between the sets and the and percentage of Exact Matches (EM).}
\label{tab:keywords_difference}
\end{table}

\begin{table}[t]
\centering
\footnotesize
\begin{tabular}{@{}r|cccc@{}}
\toprule
\multicolumn{1}{r|}{} & \geo  & \rsq  & \pen  & \nat  \\ \midrule
$r=0.5$              & 48.4 & \textbf{40.0} & 41.9 & 42.7 \\
$r=1.0 $               & \textbf{49.9} & 39.6 & \textbf{44.8} & \textbf{45.2} \\
$r=2.0$                & 49.0 & 39.4 & 41.9 & 42.0 \\ \bottomrule
\end{tabular}
\caption{Comparison of various levels of keyword intensity. We experiment with different level of keyword intensity for different oracle and train our {\aos} model on these setting. We show the F$_1$ of model's prediction and human annotation. The larger the $r$, the more keywords will be concatenated.  }
\label{tab:intensity}
\end{table}

\subsection{Ablations and Analysis}

\paragraph{Keyword Sensitivity} We evaluate the sensitivity of the model to different keywords. There is some overlap between the summaries returned by different keyword sets, as shown by the Jaccard similarity: some sentences may fit under both \geo{} and \rsq{}, or both \pen{} and \nat{}. Table~\ref{tab:keywords_difference} shows statistics of this, with the Fraud keyword sets yielding more similar summaries than those in Earthquake.
We also confirm that using the keywords ``matched'' to our setting outperforms using other sets of keywords in that domain (Table ~\ref{tab:kw-sens-domain}) suggesting that our model is picking summaries in a keyword-driven fashion. 

\paragraph{Keyword Intensity} 
We can vary the parameter $k$ controlling the number of times we append the keywords to the reference summary in order to generate the oracle extractive summary. We experiment with different level of intensity and show the result in Table~\ref{tab:intensity}. For most cases, $r=1$ works  well among all the datasets.

\section{Qualitative Evaluation \& Comparison}

\paragraph{Extractive vs. Abstractive Comparison} It is difficult to directly compare the quality of summaries produced by an extractive model to those produced by an abstractive model. Abstractive models do not  extract individual sentences from a summary so direct F$_1$ evaluations cannot be compared in the manner of Table~\ref{tab:performance}. ROUGE scores are a misleading comparison given that an extractive model will be better matched to our extractive ground truths. Therefore, we perform a qualitative analysis to determine the models' relative responsiveness to keywords and relative advantages and disadvantages.\footnote{Note that for the abstractive SPACE dataset we considered here, we found that the performance difference between our model and abstractive models is small. Our investigation found that, at least on this dataset, abstractive models are engaging in heavy copying of the source text, suggesting that extractive models may be almost as well suited for this task as abstractive models.}

\definecolor{err}{RGB}{248, 151, 31}

\begin{table}[t]
\centering
\scriptsize
\begin{tabular}{@{}cp{6.4cm}@{}} \toprule

  Sel &  \multicolumn{1}{c}{Article}\\\midrule

 & (CNN) -- A 7.2-magnitude earthquake has struck south of the Mariana Islands, according to the U.S. Geological Survey. \\

 \textit{G} & The Mariana Islands -- an archipelago in the western Pacific Ocean -- are made up of two U.S. territories, Guam and the Commonwealth of the Northern Mariana Islands.  \\
 & The islands sit about three-quarters of the way from Hawaii to the Philippines. \\

  \textit{R} & The Pacific Tsunami Warning Center did not issue a tsunami warning after \underline{the quake}, which struck at 7:19 a.m. Saturday (5:19 p.m. ET Friday).\\

 \textit{R} & "We wouldn't expect any kind of significant tsunami for this event," said the center's director, Charles McCreery, noting that the quake's magnitude was relatively low to provoke one. \\

  \textit{R} & There were no immediate reports of casualties or damage, emergency management officials said. \\

  \textit{G} & The quake struck about 375 kilometers (233 miles) west-southwest of Hagatna, Guam, and 445 kilometers (276 miles) west-southwest of Rota, Northern Mariana Islands. \\
  \midrule
\end{tabular}
\begin{tabular}{p{7cm}}
\textbf{\textsc{CTRLSum} {\geo} } \\ NEW: The location of the quake is in the province of \textcolor{err}{Yucatán}. NEW: There are no immediate reports of casualties or damage. The quake is centered about 375 kilometers (233 miles) west-southwest of Hagatna, Guam. The U.S. Geological Survey says it was a 7.2-magnitude quake. \textcolor{err}{The earthquake is centered in the Yucatan province of Mexico.} \textcolor{err}{The country's geography is similar to that of the U.N. region.}\\ \midrule
\textbf{\textsc{CTRLSum} {\rsq} } \\
NEW: The death toll from the quake is not immediately known. The U.S. Geological Survey reports a 7.2-magnitude quake. The Mariana Islands sit about three-quarters of the way from Hawaii to the Philippines. \textcolor{err}{``There is a survivor. There is an injury. There will be an aid.recovery. process,'' the U.N. secretary-general says.} The quake is centered about 375 kilometers (233 miles) west-southwest of Hagatna, Guam.\\
\bottomrule
\end{tabular}
\caption{An example article from the earthquakes domain, along with summaries selected by \aos{} (denoted as \textit{G} and \textit{R}) and \textsc{CTRLSum} with {\geo} and {\rsq} keyword. }
\label{fig:mariana}
\end{table}

\paragraph{Keyword Sensitivity Comparison}
Although both \textsc{CTRLSum} and {\aos} are sensitive to the choice of keywords and alter their summary in response to different keywords, \textsc{CTRLSum} often either hallucinates false information \cite{MaynezFactuality} or simply rewords the prompt in the generated summary. We found that just under the \geo{} keywords in the earthquakes domain, out of 100 sample articles the bigram ``not known'' appears 27 times in relation to describing the location of the earthquake and ``not immediately known'' appears another 24 times. The \textsc{CTRLSum} model frequently rephrases the prompt rather than synthesizing information in the document related to the keywords into a cogent summary.

\paragraph{Comparison of Factuality of Output} Table~\ref{fig:mariana} shows one example of \textsc{CTRLSum} hallucination in the \geo{} case. Here, the model also rewords the prompt and inserts it into the summary without adding new information. Although such behavior may possibly perform well on automated metrics, it does not serve the purpose of query-focused summarization.

\paragraph{Extractive summaries} Table~\ref{fig:mariana} shows that our model is able to successfully extract relevant parts of the document for our aspects under consideration. There are some features which may make these summaries hard to process in isolation, such as \underline{the quake} in the first \textit{R} sentence; our method could be extended with prior techniques to account for anaphora resolution \cite{durrett-etal-2016-learning}.

\section{Conclusion}
In this paper, we present a new dataset for aspect-oriented summarization of news articles called \textsc{AspectNews}. Unlike query-focused summarization datasets which are often driven by document specific facts or knowledge, this aspect-oriented task is designed to mimic common user intents in domain-specific settings. We present a keyword-controllable system trained on synthetic data and show that it can perform well on \textsc{AspectNews} without training on the target domains, performing better than a range of strong baseline methods.

\section*{Acknowledgments}

This work was chiefly supported by funding from Walmart Labs and partially supported by NSF Grant IIS-1814522, a gift from Amazon, and a gift from Salesforce Inc. Opinions expressed in this paper do not necessarily reflect the views of these sponsors. Thanks to Ido Dagan for helpful discussion and suggestions about this paper, as well to the anonymous reviewers for their thoughtful comments.

\bibliography{anthology}
\bibliographystyle{acl_natbib}

\clearpage

\appendix

\section{Appendices}
\label{sec:appendix}

\begin{figure}[t]
\centering
\includegraphics[width=0.48\textwidth]{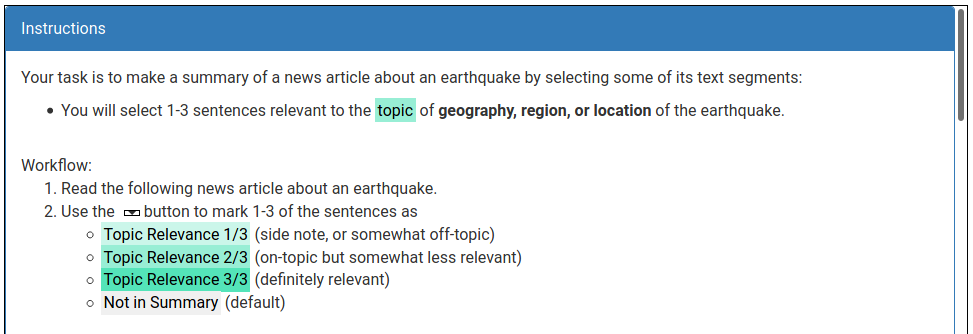}
\includegraphics[width=0.48\textwidth]{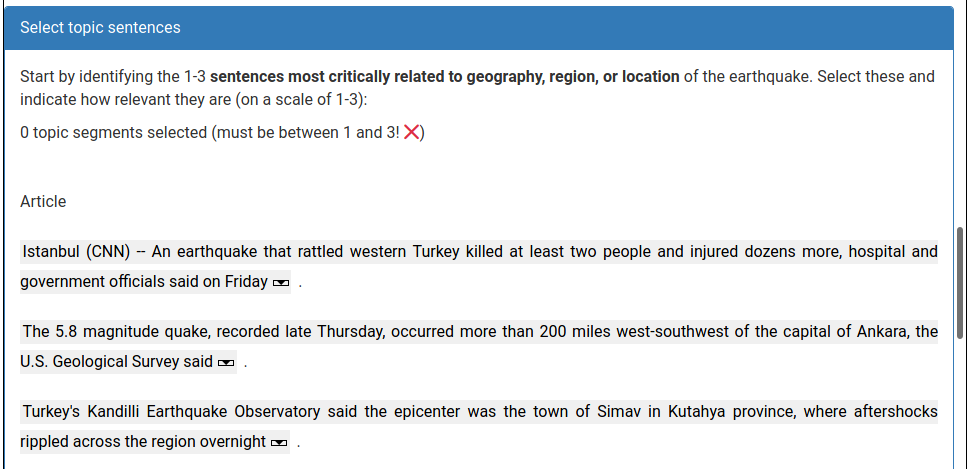}
\caption{User interface for Turkers' annotation. }
    \label{fig:turker_annotation_interface}
\end{figure}

\subsection{Training Details}
For all models, we split CNN/Daily Mail set into the standard 287,226 training pairs, 13,368 validation pairs and 11,490 test pairs following \citet{see-etal-2017-get}. 

We follow the training procedure for \textsc{BertSum} \cite{liu-lapata-2019-text} with  modifications. We use the cased variant of \texttt{bert-base-cased} available through HuggingFace \citep{Wolf2019HuggingFacesTS} instead of uncased and do not lowercase the dataset during preparation. 
Our learning rate schedule follows \citet{vaswani-2017-transformer} with $$lr = 2e^{-3} \cdot \textrm{min}(\textrm{step}^{-0.5}, \textrm{step} \cdot \textrm{warmup}^{-1.5})$$ where $\textrm{warmup} = 10000$. 

For fine-tuning \aos{} on the modified CNN/DM dataset, the training completes in 8 hours on a single NVIDIA Quadro RTX 8000.

\subsection{Exemplar Sentences}
\label{app:exemp}
In order to generate earthquake and fraud domain data we filter the CNN/DM dataset using similarity between latent representations of Universal Sentence Encoder (USE) \cite{Cer2018UniversalSE}. To find domain-related articles, we need to generate a sentence that is vague enough to match most in-domain articles but specific enough to exclude articles outside the domain. For earthquakes we found the sentence ``\emph{An earthquake occurred.}'' to work well. We embedded this sentence with USE, and calculated distance in latent space to articles in CNN/DM. For the fraud dataset we use the simlar sentence ``\emph{A fraud occured.}'' After inspecting the matches, we manually exclude articles that are outside the domain.

\subsection{Crowdsourcing}

To improve the quality of the data collected, we educate annotators with detailed instruction and user-friendly interface shown in Figure~\ref{fig:turker_annotation_interface}. We also manually sample and check the collected data.

\subsection{Oracle Derivation: BERTScore vs. ROUGE } 
\begin{table}[t]
\centering
\footnotesize
\setlength{\tabcolsep}{5pt}
\begin{tabular}{@{}r|cccc|cccc@{}}
\toprule

\multicolumn{1}{l|}{} & F$_{1}$ & R-1   & R-2   & R-L   & F$_{1}$ & R-1   & R-2   & R-L   \\ \midrule

                     & \multicolumn{4}{c|}{\pen}     & \multicolumn{4}{c}{\nat}        \\
                     RS             & 36.3 &  55.8  &  42.1  &  43.0  &  38.0  & 57.6   & 44.8  &  43.3 \\
BS             & \textbf{44.8}  & \textbf{64.2} &  \textbf{54.1}  &  \textbf{51.6} & \textbf{45.2}   &  \textbf{64.4} &   \textbf{53.9}  &  \textbf{48.0} \\
 \midrule
 
 & \multicolumn{4}{c|}{\geo}     & \multicolumn{4}{c}{\rsq}     \\
 RS             & 39.5 & 59.2  &  49.1   & 47.2 & 34.9 & 54.9  & 44.3  &  45.2 \\ 
BS             & \textbf{49.9}  & \textbf{69.1} &   \textbf{61.2} &  \textbf{54.2} & \textbf{39.6}  & \textbf{59.5} &   \textbf{49.1}  &  \textbf{46.7}   \\

\bottomrule

\end{tabular}%
\caption{Comparison of our \aos{} model trained on data using ROUGE (RS) or BERTScore (BS) as the scoring metric for oracle extraction. Training with BERTScore oracle summaries gives much stronger performance.}
\label{tab:bertscore_ablation}
\end{table}

In Table~\ref{tab:bertscore_ablation} we show the performance improvement from replacing ROUGE-derived oracle labels with their BERTScore-derived counterparts. Using BERTScore \cite{zhang-bertscore} to obtain oracle extractive summaries for training data produces models that are significantly stronger than models trained on sentences selected by maximizing ROUGE score. We hypothesize this is because ROUGE score maximization essentially limits what the model learns to lexical matching, while BERTScore can score based on more abstract, semantic criteria.

\subsection{Mixed vs. Non-Mixed} 
\begin{table}[t]
\centering
\footnotesize
\begin{tabular}{@{}r|cccc|c@{}}
\toprule
\multicolumn{1}{l|}{} & \geo  & \rsq  & \pen  & \nat &  Avg. \\ \midrule
Non-Mixed              & 48.0 & \textbf{41.8} & 43.9 & 43.7 & 44.3 \\
Mixed         & \textbf{49.9} & 39.6 & \textbf{44.8} & \textbf{45.2} & \textbf{44.9}  \\ \bottomrule
\end{tabular}
\caption{Comparison of {\aos} with or without mixed training data. We show the F$_{1}$ of the system output and human annotation on four domains. }
\label{tab:contra}
\end{table}

We compare models trained using the mixed technique against models trained without any augmentation, and find that the mixed technique generally provides some benefit, but inconsistently. In Table~\ref{tab:contra}, the Mixed technique is effective on {\geo}, {\pen}, and {\nat}, but not \textsc{Recv}. The small performance improvement from Mixed training may result from the model more easily learning the relationship between the keywords and the aspect-oriented summaries due to mixed examples. Another benefit of this technique is that a single model is capable of producing both generic and aspect-oriented summaries.

\subsection{\textsc{SPACE} Evaluation Details}
Several adjustments were made in order to run our model on the \textsc{SPACE} dataset. Since there are multiple input documents per summary, we first concatenated all documents together and treated the result as a single article. In order to process this large ``article'' with our model, we processed it in 512-token chunks using BERT in order to obtain representations from the \texttt{[CLS]} token, and then concatenated those representations together before passing them through the classification layer. This allowed selection of any sentence from any part of the input.  The following keywords were used for each of the aspects in the dataset:
\begin{enumerate*}[label=(\roman*)]
    \item service, customer, staff, employee, assistance;
    \item location, room, region, hotel, place;
    \item food, dining, restaurant, dinner, meal;
    \item building, establishment, room, property, site;
    \item cleanliness, sanitary, polished, clean, washed;
    \item rooms, chair, table, bed, wall.
\end{enumerate*}

\end{document}